\documentclass[letterpaper,10pt]{article} 
\usepackage{osameet3} 
\usepackage{devanagari}
\usepackage{pgfplotstable}
\usepackage{array}
\usepackage[mathletters]{ucs}

\usepackage{ragged2e}
\usepackage[toc,page]{appendix}
\usepackage{graphicx}
\graphicspath{ {./images} }
\usepackage{amsmath,amssymb}
\usepackage{fancyhdr}
\usepackage{lastpage}
\usepackage{float}
\usepackage{scrextend}
\begin{document}

\title{Typography-MNIST (TMNIST): an MNIST-Style Image Dataset to Categorize Glyphs and Font-Styles}
\author{
\begin{tabular}[t]{c@{\extracolsep{8em}}c} 
 Nimish Magre &  Nicholas Brown\\
Northeastern University & Northeastern University \\ 
magre.n@northeastern.edu & ni.brown@neu.edu
\end{tabular}
}

\copyrightyear{2022}

\begin{center}
\section*{Abstract}
\end{center}
\begin{center}
We present Typography-MNIST (TMNIST), a dataset comprising of 565,292 MNIST-style grayscale images representing 1,812 unique glyphs in varied styles of 1,355 \textit{Google-fonts}\footnote{https://fonts.google.com/\label{1}}.The glyph-list contains common characters from over 150 of the modern and historical language scripts with symbol sets, and each font-style represents varying subsets of the total unique glyphs. The dataset has been developed as part of the \textit{Cognitive Type} project ~\cite{Cognitive-Type} which aims to develop eye-tracking tools for real-time mapping of type to cognition and to create computational tools that allow for the easy design of typefaces with cognitive properties such as readability. The dataset and scripts to generate MNIST-style images for glyphs in different font styles are freely available at \textit{https://github.com/aiskunks/CognitiveType}.    
\end{center}

\section{Introduction}

Since its introduction in 1998 by LeCun et al.~\cite{lecun1998gradient}, the MNIST dataset, comprising of handwritten digit images for classes 0-9, has become increasingly popular with deep learning researchers. Researchers are able to test their algorithms on this real-world data relatively quicker due to its small sized samples and minimal pre processing requirements. This is precisely the reason behind producing the TMNIST dataset with pre processing steps that mirror the MNIST dataset. Therefore, each grayscale image produced in the TMNIST dataset has a standard $28\times28$ size with the glyph image centered and sized to $20\times20$.

\noindent
With this work, we hope to provide researchers with a practical dataset for performing font and glyph classification tasks. In comparison to both MNIST and EMNIST~\cite{7966217} (an extended version of MNIST with lower and uppercase Latin alphabets), TMNIST could prove to be a more challenging classification dataset due to the introduction of a higher number of classes in terms of both glyph-labels and font-styles.

\section {Typography-MNIST Dataset}
The Typography-MNIST dataset relies on the freely accessible binary \textit{Google-font} files \footnote[2]{https://github.com/google/fonts} and a combination of all unique sample-glyphs represented by each of the 1,355 \textit{Google fonts}\footref{1}. Since \textit{Google} forbids web-scraping from the \textit{Google-fonts} subsite, and a comprehensive list of all possible glyphs that can be configured by each of the \textit{Google-fonts} is unavailable, a list of concatenated sample glyphs configured by each of the 1,355 \textit{Google-fonts} was manually attained from the \textit{Google-fonts} site \footref{1}. Using this list, a unique set of 1,819 glyphs was compiled and is available along with the dataset.\\
The following conversion pipeline was then used to generate MNIST style images from the list of unique glyphs and binary font files:
\begin{enumerate}
    \item import the \textit{.ttf}/\textit{.otf} binary font file
    \item verify if the glyph can be configured by the particular font file
    \item display the glyph in the particular font-style on a blank white canvas with font-size 28 and font-color 'black'
    \item convert the image to gray-scale and invert image pixel values
    \item crop the glyph portion of the image and resize to $20\times20$ without losing aspect ratio
    \item add zero padding to resize the image to $28\times28$
    \item compute the weighted average (center of mass) of the pixel intensities and recenter this point to the center of the $28\times28$ field
\end{enumerate}

\begin{figure}[ht]
  \centering
  \includegraphics{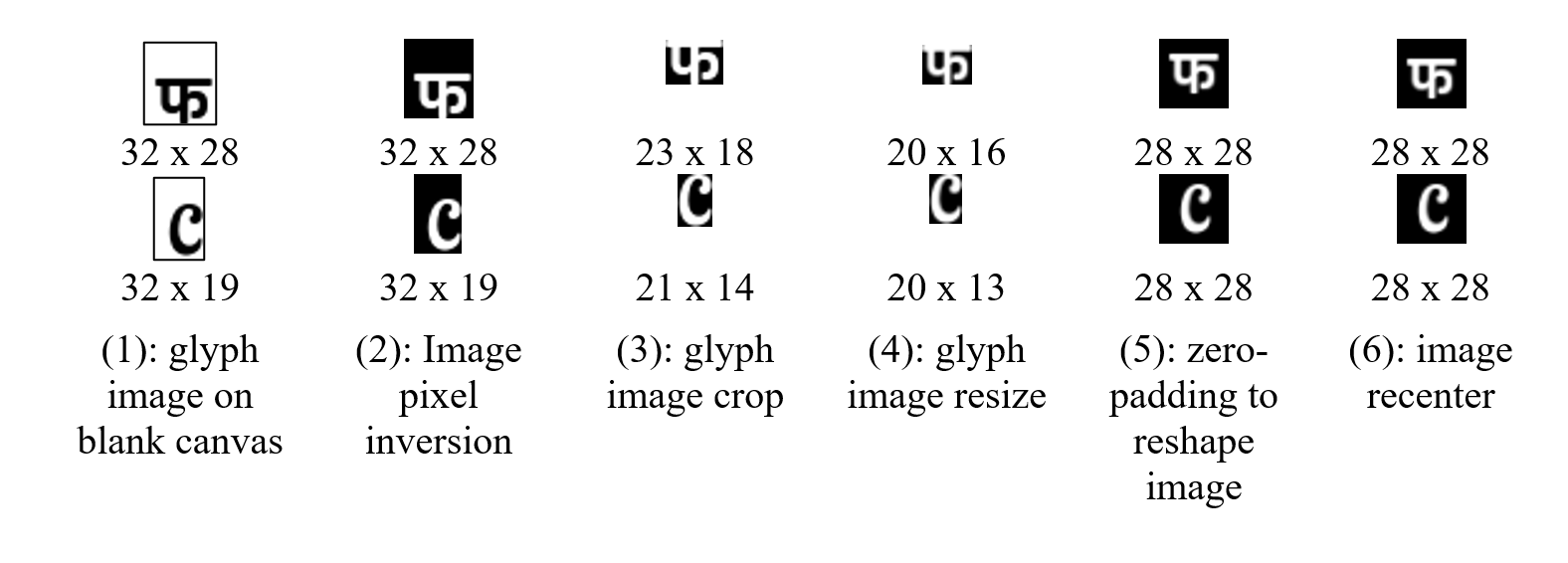}
\caption{Visualization of the process used to generate TMNIST data. The two examples depicted display the \textit{Devanagari Letter Pha} in the \textit{Rajdhani-Bold} font-style and the \textit{Latin Capital Letter C} in the \textit{Aladin-Regular} font-style respectively.}
\end{figure}

\noindent
The dataset is stored in a \textit{csv} file where the first column represents the Font-style, the second column represents the Glyph-label and the remaining 786 columns represent the $28\times28$ gray-scale image pixel values. The Font-style (ex: \textit{aladin-regular, abel-Bold}) is obtained from the binary font-file name whereas the Glyph-label (ex: \textit{LATIN CAPITAL LETTER C, DEVANAGARI LETTER PHA}) is obtained through the \textit{Unicode Character Database}\footnote[3]{https://www.unicode.org/reports/tr44/} for the specific glyph. Certain glyphs such as {\dn a\2} are represented as a combination of two \textit{Unicode data-names} and therefore the glyph label is represented as a combination of the two \textit{Unicode data-names} separated by the '+' symbol. Hence, the glyph-label for the character {\dn a\2} is (DEVANAGARI LETTER A + DEVANAGARI SIGN ANUSVARA).

\noindent
Two subsets of the dataset that contain images of digits from 0-9 only (\textit{TMNIST-Digit})\footnote[4]{https://www.kaggle.com/nimishmagre/tmnist-typeface-mnist} and images of upper and lower-case Latin alphabets (\textit{TMNIST-Alphabet})\footnote[5]{https://www.kaggle.com/nikbearbrown/tmnist-alphabet-94-characters} respectively, along with the original \textit{TMNIST dataset}\footnote[6]{https://www.kaggle.com/nimishmagre/tmnist-glyphs-1812-characters} have also been released. The promising classification results obtained on these images are available with the datasets and validate their utility.
A more informal method of randomly displaying 100 images from the dataset and manually verifying the labels was also successfully applied to validate the data. 

\noindent
The following table summarizes the TMNIST dataset files released so far:

\begin{table}[H]
\caption{Files contained in the Typography-MNIST dataset}
  \centering
  \includegraphics[scale=0.8]{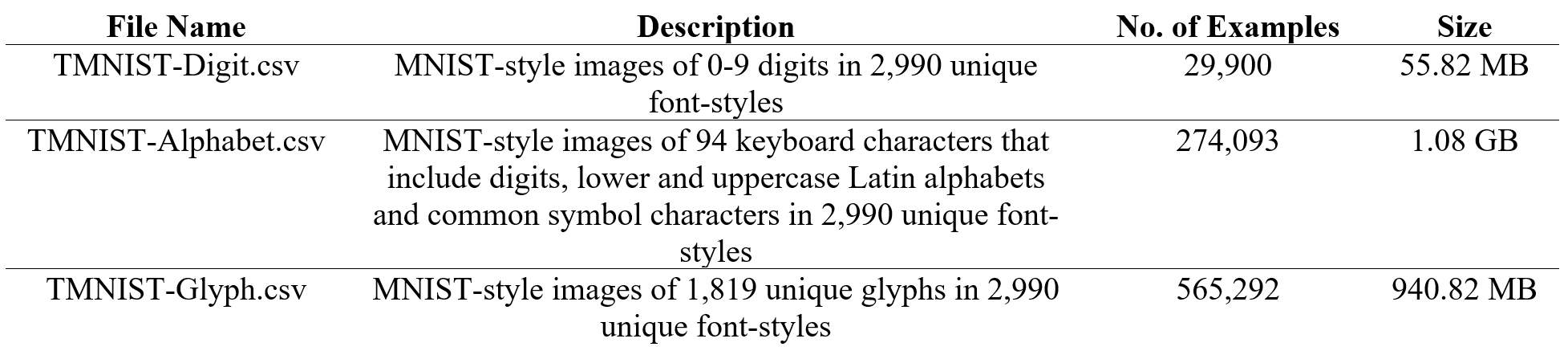}
\end{table}

\section{Conclusion}
This paper introduces Typography-MNIST (TMNIST), an MNIST-style image dataset of unique glyphs presented in varying font-styles. The dataset is intended to provide researchers the advantages of the MNIST data whilst performing glyph and font-style classification tasks. An extended version of the dataset with elastic distortions is also available for use and the users are free to augment the dataset further if required.

\bibliographystyle{plain}
\bibliography{bibliography.bib}

\end{document}